\documentclass[letterpaper, 10 pt, conference]{ieeeconf}  

\IEEEoverridecommandlockouts                              

\usepackage[l2tabu,orthodox]{nag}

\usepackage[
backend=bibtex8, 
style=ieee, 
sorting=none, 
natbib=true, 
doi=false, 
isbn=false, 
url=false, 
eprint=false, 
maxcitenames=1, 
mincitenames=1
]{biblatex}

\usepackage[draft,pdftex,colorlinks]{hyperref}

\usepackage[printonlyused]{acronym}

\usepackage{siunitx}
\usepackage[all]{nowidow}

\usepackage[dvipsnames]{xcolor}

\usepackage{lipsum}

\usepackage[pdftex]{graphicx}

\usepackage{epstopdf}

\usepackage{import}

\graphicspath{{./latexGoodPractices/}}

\usepackage{booktabs}

\usepackage{tabu}

\usepackage{amssymb,amsfonts,amsmath,amscd}

\usepackage{bm} 

\newcommand{\bbm}{\begin{bmatrix}}
\newcommand{\ebm}{\end{bmatrix}}

\usepackage{tikz}
\usepackage{graphicx}
\usepackage{subcaption}
\usepackage{commath}
\DeclareMathOperator*{\argmax}{arg\,max}

\title{\LARGE \bf
	Lidar Measurement Bias Estimation via Return Waveform Modelling in a Context of 3D Mapping 
}

\author{Johann Laconte\authorrefmark{1}\authorrefmark{2}, Simon-Pierre Desch\^{e}nes\authorrefmark{2}, Mathieu Labussi\`ere\authorrefmark{1}\authorrefmark{2}, Fran\c cois Pomerleau\authorrefmark{2}
	\thanks{\authorrefmark{1} The authors are with Institut Pascal, Campus des Cezeaux, 63178 Aubi\`{e}re Cedex, France.
     {\tt\small laconte.johann@gmail.com}}%
\thanks{\authorrefmark{2} The authors are with Northern Robotics Laboratory, Universit\'{e} Laval, Canada.
{\tt\small francois.pomerleau@ift.ulaval.ca}}%
}

\usepackage{xspace}
\acrodef{LIDAR}{Light Detection And Ranging}
\acrodefplural{LIDAR}{Light Detection And Ranging}
\newcommand{\LMS}{LMS151\xspace}
\newcommand{\HDL}{HDL-32E\xspace}
\newcommand{\RS}{RS-LiDAR-16\xspace}
\newcommand{\ie}{i.e., }

\newcommand{\expp}[1]{\exp\left(#1\right)}
\newcommand{\epp}[1]{\mathrm{P}_{\text{e}}\left(#1\right)}
\newcommand{\ppp}[1]{\mathrm{P}_{\text{p}}\left(#1\right)}
\newcommand{\rpp}[1]{\mathrm{P}_{\text{r}}\left(#1\right)}
\DeclareMathOperator{\erf}{erf}

\begin{document}
\usetikzlibrary{shapes.geometric, arrows}

\maketitle
\thispagestyle{empty}
\pagestyle{empty}

\begin{abstract}
	In a context of 3D mapping, it is very important to obtain accurate measurements from sensors.
	In particular, \ac{LIDAR} measurements are typically treated as a zero-mean Gaussian distribution.
	We show that this assumption leads to predictable localisation drifts, especially when a bias related to measuring obstacles with high incidence angles is not taken into consideration.
	Moreover, we present a way to physically understand and model this bias, which generalizes to multiple sensors.
	Using an experimental setup, we measured the bias of the Sick \LMS, Velodyne \HDL, and Robosense \RS as a function of depth and incidence angle, and showed that the bias can reach \SI[detect-weight=true]{20}{\cm} for high incidence angles.
	We then used our model to remove the bias from the measurements, leading to more accurate maps and a reduced localisation drift.
\end{abstract}
\begin{keywords}
	Bias Estimation, Sensor Error Modelling, Waveform Modelling, LIDAR, 3D Mapping
\end{keywords}



\section{Introduction}
\acp{LIDAR} have become widespread in robotics to produce increasingly precise maps of the environment.
%
As every sensor, \acp{LIDAR} suffer from noise, directly impacting the quality of maps. 
In particular, we observed that the incidence angle of the laser beam can cause significant errors in mapping algorithms.

\begin{figure}[htb]
	\centering
	\includegraphics[width=\columnwidth]{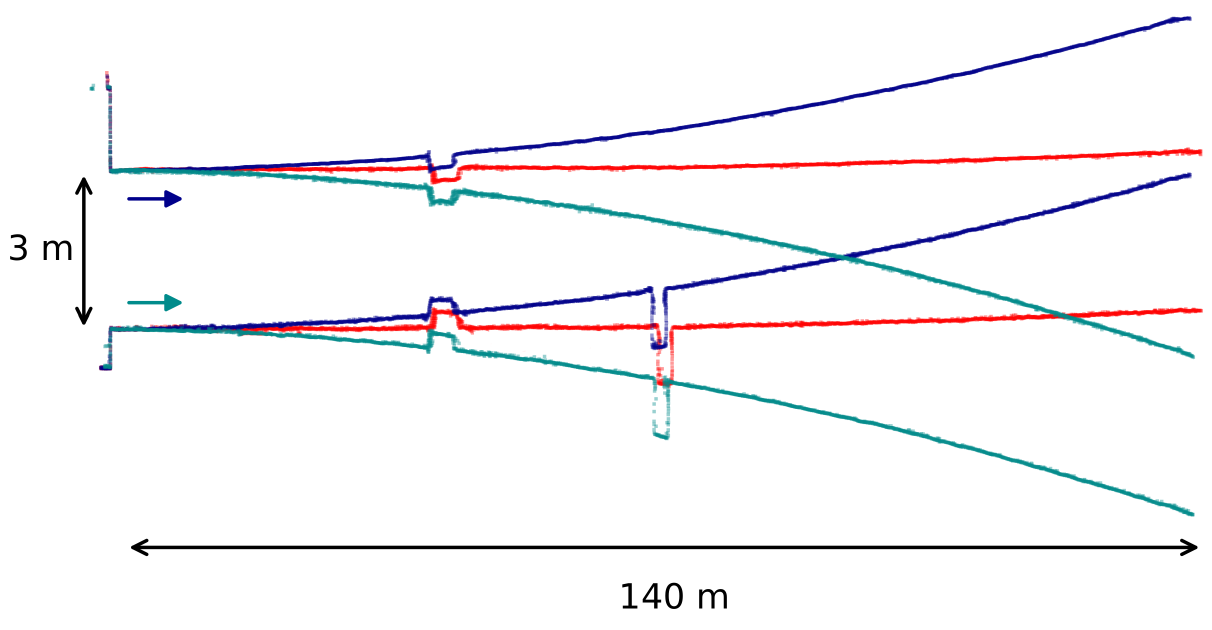}
	\caption{Impact of bias from \ac{LIDAR} measurements on mapping.
		The map is bending depending on which wall the robot was closest to while driving.
		Coloured arrows represent the starting position of the robot and the wall it followed (blue: left; cyan: right).
		Once the bias is taken into consideration, the map (red) drifts significantly less.
		Note that the scale on the length of the corridor was changed to highlight the impact of the bias.
		The corridor is 47 times longer than its width.
	} 
	\label{fig:twisted_map}
	\vspace{-2mm}
\end{figure}

For example, when a map of the environment is produced in a tunnel or a hallway, notable errors occur in the distance where the incidence angle is very important.
The error can be a simple zero-mean noise.
However, it is possible that the noise is not truly centred on zero, hence has a bias. 
\autoref{fig:twisted_map} shows a map taken from a straight underground tunnel, produced with the \LMS.
Although the tunnel is known to be straight, the produced map tends to bend at large distances. 
Moreover, the tunnel bends in a preferred direction depending on which wall the \ac{LIDAR} is closest.
This deformation is repeatable given a trajectory, which disproves the idea that the bending is only induced by the integration of a zero-mean noise.
This observation leads to the hypothesis that \acp{LIDAR} have a bias that depends on the distance and the incidence angle.
Modelling this bias would correct the measurements and improve the map quality.
\autoref{fig:twisted_map} shows the unbent map, where the bias was removed from the \ac{LIDAR} measurements.

The contribution of this paper is a physical explanation for the bias, as well as a way to quantify and correct this bias for common \acp{LIDAR} used in mobile robotics.
\autoref{section:RelatedWork} presents a survey of the modelisation of the noise of \acp{LIDAR}, often considered as a zero-mean white noise. 
\autoref{section:Simulation} describes the modelisation of the return waveform.
We test our hypothesis in \autoref{section:ExperimentalSetup} with an experimental setup for the \acp{LIDAR} Sick \LMS, Velodyne \HDL, and the new Robosense \RS(characterised in \cite{Wang2017}), leading to a novel method to correct the measurements of \acp{LIDAR} whose results are presented in \autoref{section:Results}.

\section{Related Works} \label{section:RelatedWork}
In the literature, many have shown the relationship between the noise of \ac{LIDAR} measurements and the incidence angle. 
An extensive study of the influence of the incidence angle on the measurements was conducted in~\citep{Soudarissanane2007,Soudarissanane2009,Soudarissanane2011}.
The authors focused on the propagation of the noise induced by the incidence angle, albeit assuming no bias. 
\citet{Gombak2016} created a probabilistic model of \acp{LIDAR} and found that the behaviour of the Hokuyo UTM-30LX at high incidence angles is different from normal distributions.
They discovered a bias at high incidence angles but mainly because the beam was close to a corner, hence measuring the reflexivity of the laser beam and not the bias itself. 
\citet{Pomerleau2012} analysed the noise from several surfaces and stated that standard models for \ac{LIDAR} (\ie without bias) overestimate the variance to cope with unmodelled biases.
Our hypothesis is reinforced by~\citet{Pfister}, who characterised the bias of the Sick LMS-200 as a function of the incidence angle and the depth and then integrated it to their mapping algorithm.
They modelled the bias using an empirical function and concluded that the contribution of the correction is negligible in their environment, which had many features and landmarks.
In this paper, we demonstrate that the contribution of the bias for the 2D and 3D cases is not negligible in all environments.

\acp{LIDAR} were characterised as a function of many factors, such as depth, internal temperature or incidence angle.
\citet{Reina1997} showed that the target seems to move away from the sensor as the incidence angle increases. 
\citet{CangYe2002} characterised the Sick LMS200 using an experimental setup similar to the one used in the present article and concluded that the incidence angle has a great influence on the reading. 
They did not model the error as a function of the incidence angle as they stated that this information is not usually known in most mobile robotics applications.
It is nowadays possible to estimate this parameter, which is done in this article.
The same experimental setup was used by~\citet{Okubo2009} to characterise the Hokuyo URG-04LX, but the authors were not able to obtain reliable measurements for an incidence angle greater than \SI{40}{\degree}. 
They therefore could not observe the bias, which manifests itself at high incidence angles.
\citet{Kidd2016} studied the performances of the Velodyne VLP-16 at different incidence angles, highlighting the bias induced by it.
They gave a partial estimate of the bias but were unable to connect it with a physical phenomenon, leading to large unmodelled errors.
At a larger scale,~\citet{Roca-Pardinas2014} conducted an analysis of the influence of terrestrial laser scanning for tunnel inspection, and modelled the error as a function of the incidence angle and the depth with a non-parametric kernel smoother. Nevertheless, this provides no physical explanation of the bias. 
This last solution is sensitive to the density of the experimental characterisation, which could be simplified using an analytic solution.

\acp{LIDAR} were also modelled and integrated in simulation softwares. 
\citet{Al-Temeemy2017} and \citet{Kruapech2010b} modelled a \ac{LIDAR} return waveform using a Gaussian beam representation to compute the laser beam propagation. They however did not consider the incidence angle.
Our work extends these simulations by considering it. 
\citet{Budge} modelled further the return waveform by taking into account the electronics of the sensor and the discrimination method used to find the maximum of the return waveform.
Our method differs from the last one as~\cite{Budge} discretises the laser beam, mainly to handle the mixed pixel problem~\cite{Hebert}.
Finally,~\citet{Wagner2004b} concluded that the study of the full return waveform leads to better results than using an electronic peak detector to estimate the distance. 
This statement supports our theory that the current detectors embedded in \acp{LIDAR} cannot handle all of the situations the beams can encounter.

\section{Modelling of the Waveform} \label{section:Simulation}

\begin{figure}[t!]
	\centering
	\includegraphics{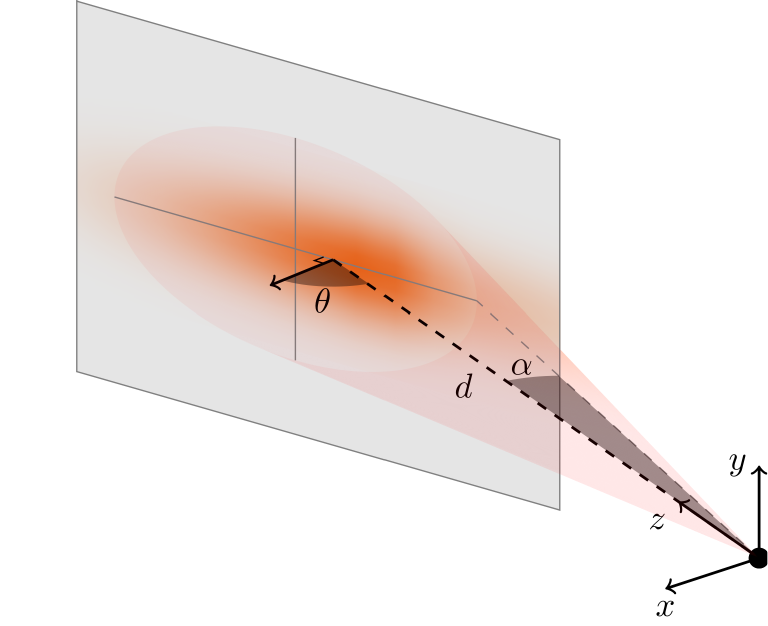}
	\caption{Illustration of a single laser beam (shaded in red), with an aperture half-angle $\alpha$, hitting an oriented surface of $\theta$ degrees at a distance $d$.
	The bias comes from the intensity distribution being skewed causing the peak of intensity to be detected sooner once integrated by the receptor.
	}
	\label{fig:projectedBeam}
\end{figure}

A first intuition of the phenomenon is illustrated in \autoref{fig:projectedBeam}.
When the laser beam hits an oriented surface, the energy distribution is not uniform. 
A great amount of intensity returns sooner to the sensor than if the surface was not oriented. 
This means that the return waveform will have a different shape depending on the incidence angle, inducing a bias in the measurement. 

We simulate the return waveform of a \ac{LIDAR} using a four-step process.
(1) A pulse of light is emitted, going through the lens of the \ac{LIDAR}.
(2) The beam is then projected onto the oriented plane.
(3) The beam is reflected.
(4) The light comes back to the \ac{LIDAR} as a function of time, taking into account that the travel time for a given point on the plane differs according to its position.

\begin{figure}[t!]
	\vspace{-3mm}
	\centering
	\includegraphics[width=.8\columnwidth, trim=1.0cm .6cm 0cm .3cm, clip]{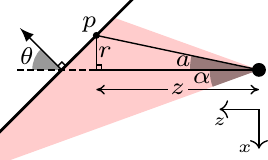}
	\caption{Notations for the 2D case. 
	Each point $p$ on the plane is defined as a function of $a$, the angle formed with the centre of the beam.
	$p$ is at an orthogonal distance $z$ from the \ac{LIDAR} and at a distance $r$ from the centre of the beam.
	}
	\label{fig:notations}
	\vspace{-4mm}
\end{figure}

Without loss of generality, we place the sensor at the origin, facing the $\vec{z}$ axis. 
The plane is parallel to the $\vec{y}$ axis and its normal forms an angle $\theta$ with the $\vec{z}$ axis. 
\autoref{fig:notations} summarises the notations for the 2D case.

We model the emission of the light pulse $\epp{t}$ with a Gaussian of length $\tau$ and power $I_0$ as done in~\cite{Budge}
\begin{equation}
	\epp{t} = I_0 \expp{{\frac{-t^2}{2\sigma^2}}},
\end{equation}
with $\sigma = \frac{\tau}{\sqrt{2\pi}}$.
%
The beam is then projected onto the plane at a distance $d$ and rotated by \SI[parse-numbers=false]{\theta}{\degree}. 
Since we cannot know the reflexivity properties of the plane, we chose to use the Lambertian model $\mathrm{L}(\cdot)$ for the projected energy $\ppp{\cdot}$, given a known plane, as 
\begin{equation} \label{eq:projected}
	\ppp{t,a,b \, \middle| \, d, \theta} = \mathrm{L}(\gamma)\cdot \epp{t-\frac{\rho}{c}} \cdot \mathrm{I}(r,z),
\end{equation}
with $\mathrm{L}(\gamma)=\cos(\gamma)$, $\gamma=\cos^{-1}(\cos(a+\theta)\cos b)$, $c$ the speed of light,
$a$ the angle between the point and the centre of the beam along the major axis of the ellipse (\ie on the plane $\vec{x}\vec{z}$), $b$ the angle between the point and the centre of the beam along the minor axis (\ie on the plane $\vec{y}\vec{z}$),
$r$ the distance from the centre of the beam and $z$ the distance of the centre of the beam from the emitter (\autoref{fig:notations}). 
The length $\rho$ corresponds to the distance between the emitter and the point and can be computed as follows
\begin{align}
	\rho&=\sqrt{
		\smash[b]{\underbrace{\vphantom{ \left(d \frac{\cos(a) \cos(\theta)}{\cos(a+\theta)}  \right)^2 } z^2\tan^2(a)+z^2\tan^2(b)}_{r^2}} 
	+ 
	\smash[b]{\underbrace{\left(d \frac{\cos a \cos \theta}{\cos(a+\theta)}  \right)^2}_{z^2}}
	}.
	\\ \nonumber 
\end{align}
\autoref{eq:projected} also uses the Gaussian beam model $\mathrm{I}(\cdot)$, with a wavelength $\lambda$ and the aperture half-angle $\alpha$, defined as 
\begin{equation}
	\mathrm{I}\left(r,z\, \middle| \, \lambda, \alpha \right) = \left(\frac{\omega_0}{\alpha z}\right)^2\cdot \expp{\frac{-2r^2}{\alpha^2z^2}},
\end{equation}
at large distance, with $\omega_0=\frac{\lambda}{\pi\alpha}$. The function $\mathrm{I}(\cdot)$ therefore represents the intensity for the points along the basis of a circular cone of radius $r$ and height $z$.

\begin{figure}[thb]
	\centering
	\includegraphics[width=\columnwidth]{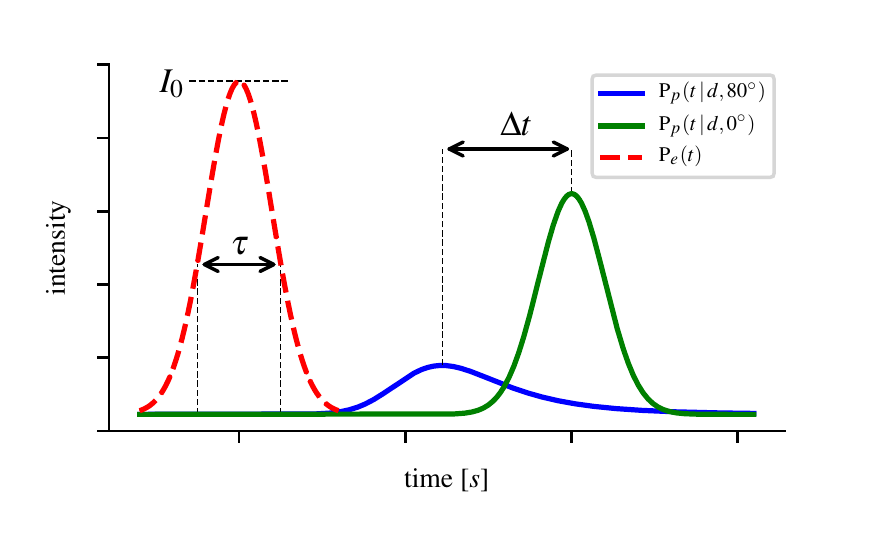}
	\vspace{-7mm}
	\caption{
	Illustration of two return waveforms hitting a surface at the same distance. 
	In red, the emitted waveform with power $I_0$ and length $\tau$.
	In green, the return waveform with $\theta=\SI{0}{\degree}$. 
	In blue, the return waveform with $\theta=\SI{80}{\degree}$.
	The peaks of intensity are spaced apart of $\SI[parse-numbers=false]{\Delta t}{\s}$. 
	The curves also differ in shape, with the waveforms flattening as the incidence angle grows. 
	}
	\label{fig:RP}
	\vspace{-2mm}
\end{figure}

The beam is then returned to the sensor.
To determine the total energy returned to the detector $\rpp{\cdot}$, we need to integrate over the surface (\ie in both directions) of the plane at each time $t$, leading to
\begin{equation}
	\rpp{t \, \middle| \, d, \theta} = \int_{-\frac{\pi}{2}}^{+\frac{\pi}{2}}\int_{-\frac{\pi}{2}}^{+\frac{\pi}{2}}  \ppp{t-\frac{\rho}{c}, a, b \, \middle| \, d, \theta} \dif b \dif a	,
	\label{eqRP}
\end{equation}
\autoref{fig:RP} gives an example of the simulation of the return waveform $\rpp{\cdot}$ for two different angles of incidence.
The shape of the return waveform tends to skew as the incidence angle grows, moving the peak of intensity by $\Delta t$ (i.e., leading to a bias in the measurement).

One can easily notice that the equation for the return waveform is not computational-friendly. 
In order to find a closed-form of \autoref{eqRP}, we restrict the simulation to the 2D case (\ie keep $b=0$), therefore having only one integral to cope with.
This approximation can be made because the distribution of intensity along the $\vec{y}$ axis is not skewed, therefore inducing no further bias (\autoref{fig:projectedBeam}).
We also bound the integral using the aperture half-angle $\alpha$. The intensity is indeed negligible outside these boundaries. 
We obtain
\begin{equation}
	\rpp{t \, \middle| \, d, \theta} \approx \int_{-\alpha}^{+\alpha} \ppp{t-\frac{\rho}{c}, a, 0 \, \middle| \, d, \theta} \dif a.
\end{equation}
%
Using first-order Taylor series, we find  
\begin{equation}
	\rpp{t\, \middle| \, d, \theta}\approx I_0 \left(\frac{w_0}{\alpha d \cos(\theta)}\right)^2 \expp{C}\int_{-\alpha}^{+\alpha} f(a)\dif a,
\end{equation}
with $f(a) =(K_1-aK_2)\exp(-Aa^2+Ba)$ and
\begin{align}
		A &= \frac{2d^2\tan^2(\theta)}{\sigma^2c^2} + \frac{2}{\alpha^2}, &K_1 &= \cos^3(\theta), \notag\\
		B &= \frac{1}{\sigma^2}\left(t-\frac{2d}{c}\right)\left(\frac{2d\tan(\theta)}{c}\right), &K_2 &= 3\cos^2(\theta)\sin(\theta), \notag\\
	C &= -\frac{1}{2\sigma^2}\left(t-\frac{2d}{c}\right)^2, &&
\end{align}
which has a closed-form.
We will now concentrate on the peak of $\rpp{t\, \middle| \, d, \theta}$, hence when $t$ is in the neighbourhood of $\frac{2d}{c}$ (where the peak should be if $\theta$ was null). 
Under this assumption, we can simplify the above equation using again first-order Taylor series. Using $T=t-\frac{2d}{c}$, we express $\rpp{\cdot}$ as 
%
\begin{equation} \label{eq:finaleq}
	\begin{aligned}
		\rpp{t\, \middle| \, d, \theta}\approx& \sum_{i=0}^3 a_iT^i,
	\end{aligned}
\end{equation}
with
\begin{align}
		a_0 &= 2AK_1L_1, \notag\\
		a_1 &= -\frac{2d\tan(\theta) \left(-2L_2\alpha \expp{-A\alpha^2}+L_1K_2\right)}{\sigma^2c}, \notag\\
		a_2 &= - \frac{2AK_1L_1 \left(\sigma^2c^2A\cos^2(\theta)+2d^2\cos^2(\theta)-2d^2\right)}{2\cos^2(\theta)\sigma^4c^2A}, \notag\\
		a_3 &= \frac{L_1K_2d\tan(\theta)\left(\sigma^2c^2A-2d^2\tan^2(\theta)\right)}{\sigma^6c^3A}, \\ 
		L_1 &= I_0 \cdot\left(\frac{w_0}{\alpha d \cos(\theta)}\right)^2 \cdot\frac{\sqrt{\pi} \erf\left(\alpha\sqrt{A}\right)}{2A^{3/2}}, \notag\\
		L_2 &= I_0 \cdot\left(\frac{w_0}{\alpha d \cos(\theta)}\right)^2 \cdot\frac{K_2}{2A}. \notag
\end{align}

At this point, we have a function depending on three variables: $d$, $\theta$ and $t$. We need to eliminate the third variable, by defining some metrics on $\rpp{t\, \middle| \, d, \theta}$.
The most logical is the difference of distance the \ac{LIDAR} would return between the waveforms at $\SI{0}{\degree}$ and $\SI[parse-numbers=false]{\theta}{\degree}$, if the electronics was able to detect without error the maximum of intensity:
\begin{equation}
	\begin{aligned}
		\Delta_d(d,\theta) &= [\argmax_t \rpp{t\, \middle| \, d, \theta}]\cdot \frac{c}{2} - d \\
						   &=\frac{-2a_2 - \sqrt{4{a_2}^2-12a_1a_3}}{6a_3} \cdot\frac{c}{2}
	\end{aligned}
\end{equation}
%

However,~\citet{Wagner2004b} showed that no detector is perfect and that an error is always made.
We therefore define another metric to measure the difference of shape between the waveforms at \SI{0}{\degree} and \SI[parse-numbers=false]{\theta}{\degree}:
\begin{equation}
	\begin{aligned}
		\Delta_{\textrm{shape}}(d, \theta) &= 1-\frac{\kappa(d,0)}{\kappa(d,\theta)}
	\end{aligned}
\end{equation}
with 
\begin{equation}
	\begin{aligned}
		\kappa(d,\theta)&= \sqrt{4{a_2}^2-12a_1a_3},
	\end{aligned}
\end{equation} 
the curvature of the return waveform at its peak.


Because of the many uncertainties on the variables and the model which we restricted to the 2D case, we allow our model to have two scaling factors $(s_1,s_2)$. The error of distance is then expressed as 
\begin{equation}
	e(d,\theta) = s_1\cdot \Delta_d(d,\theta) + s_2\cdot \Delta_{\mathrm{shape}}(d,\theta).
	\label{eq:diff}
\end{equation}
These scaling factors will be adjusted for each sensor.

\section{Experimental Setup} \label{section:ExperimentalSetup}
\begin{figure}[t!]
	\centering
	\includegraphics[width=\columnwidth]{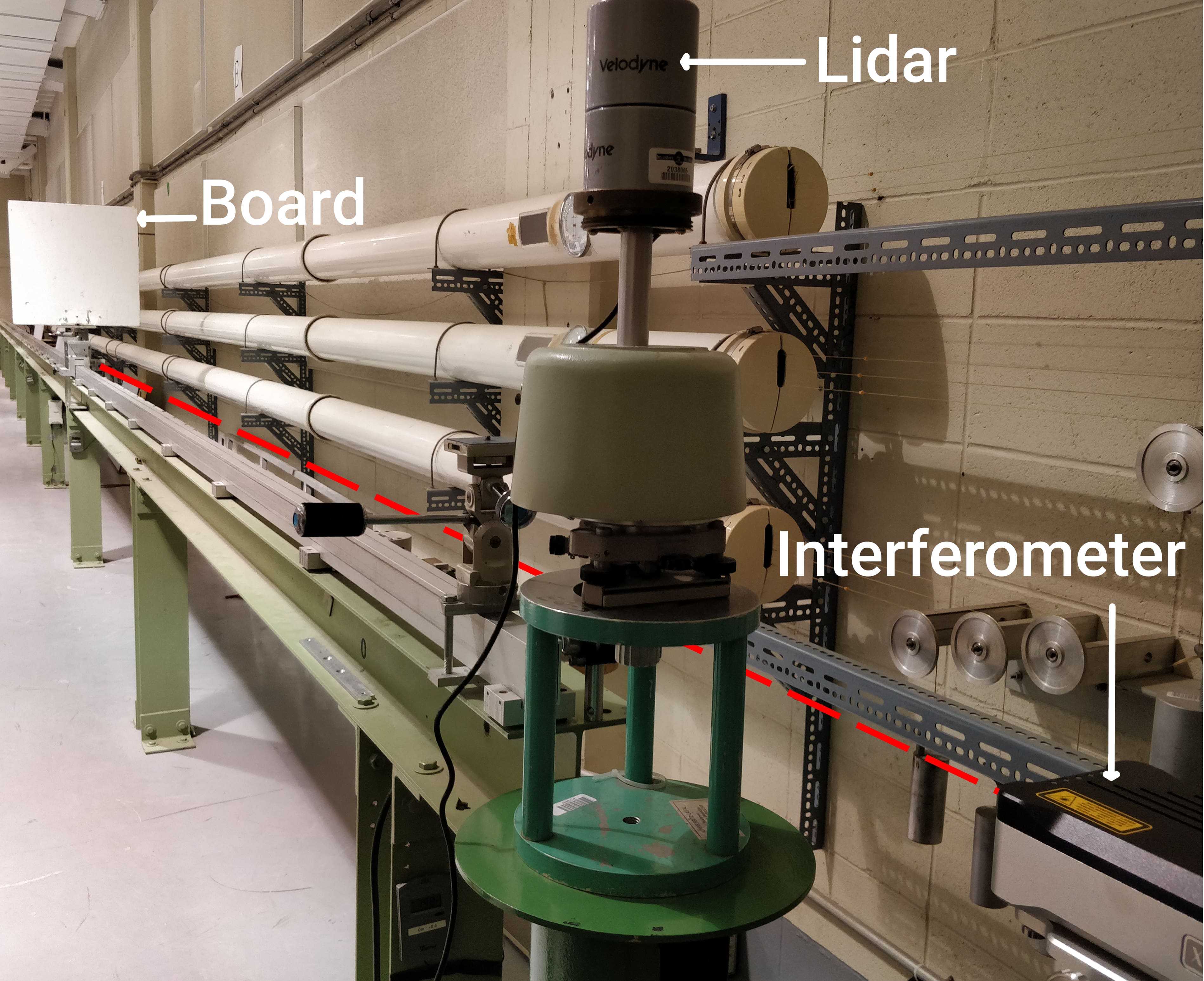}
	\caption{Experimental setup. The \ac{LIDAR} central beam measures the board which can be rotated gradually. 
	The whole setup is mounted onto a linear motion table, allowing measurements to be made at several distances. 
	The interferometer measures the distance of the board, with its beam indicated with a red dashed-line.
	}
	\label{fig:setupPic}
\end{figure}

In order to test our hypothesis, we used an experimental setup very similar to the one in~\cite{CangYe2002}, as shown in \autoref{fig:setupPic}.
A wooden board of \SI[parse-numbers=false]{0.6 \times 0.6}{\meter} was mounted onto a linear motion table (\ie a rail) of approximately \SI{30}{\meter}. 
The board can rotate using a graduated disc.

An interferometer (Renishaw XL-80) gives the ground truth distance $D$ between the board and the \ac{LIDAR}, with an accuracy of the order of \SI{1}{\um}. 
The \ac{LIDAR} itself is rigidly fixed with respect to the linear motion table with its centre beam aligned with the rail, measuring a distance $d$.

\begin{figure}[t!]
	\centering
	\includegraphics[width=\columnwidth]{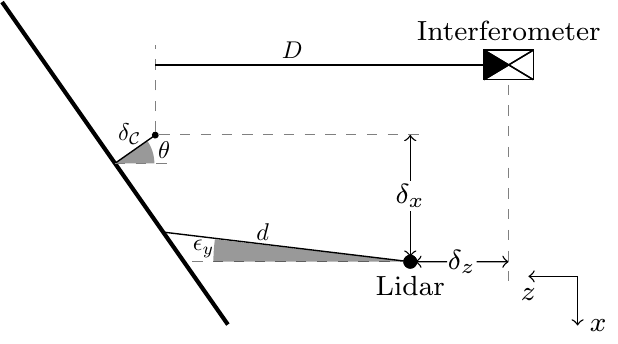}
	\caption{Top-view of the experimental setup. The \ac{LIDAR} measures the board at a distance $d$ and is misaligned by a small rotation of $\epsilon_y$. 
		It is also misplaced from the rail by $\delta_x$. The board rotates at a distance $\delta_\mathcal{C}$ from its centre of rotation.
	The interferometer is at a distance $\delta_z$ of the \ac{LIDAR} and measures a distance $D$.}
	\label{fig:setupErrors}
\end{figure}

As it can be seen in \autoref{fig:setupErrors}, several parameters need to be adjusted before any measurement. 
First, the error of orientation $\epsilon_y$ of the sensor was cancelled by aiming the \ac{LIDAR} beam at a very thin object (\SI{5}{\cm}) on the linear motion table at a large distance (\ie more than \SI{30}{\meter}).
Second, the displacement $\delta_z$ between the interferometer and the \ac{LIDAR} was computed as the mean of the difference of distances returned by the \ac{LIDAR} and the interferometer with the board at \SI{0}{\degree}.
Third, we manually measured the distance $\delta_\mathcal{C}$ between the board and its centre of rotation.
Finally, we computed the horizontal displacement $\delta_x$ by measuring with the \ac{LIDAR} the distance to the board at different orientations (all below \SI{30}{\degree}, where the bias is assumed to be negligible) and estimated the centre of rotation of the board.

Even if we did our best to minimise these errors, it is very likely that some parameters are not precise enough and could interfere with the measurements. 
The most important parameter is the rotation $\epsilon_y$ of the sensor, which could lead to great errors at high angles and compensate or enhance the studied phenomenon.
As validated by~\citet{CangYe2002}, we took the mean of the errors at $\SI[parse-numbers=false]{\theta}{\degree}$ and $\SI[parse-numbers=false]{-\theta}{\degree}$ to minimise the influence of $\epsilon_y$ and $\delta_x$.

Knowing all the parameters, for a distance of measurement $d$ and an incidence angle $\theta$, the corrected distance that the \ac{LIDAR} should return is
\begin{equation}
	D_c= D - \delta_z + \frac{\delta_\mathcal{C}}{\cos(\theta)} - \delta_x\tan(\theta).
\end{equation}
The error (\ie bias) is computed as
\begin{equation}
	e(d,\theta) = d- D_c.
\end{equation}
%
%

For all three \acp{LIDAR}, we measured the distance error for twelve orientations of the board.
We used steps of \SI{10}{\degree} from \SI{0}{\degree} to \SI{60}{\degree} and steps of \SI{5}{\degree} from \SI{65}{\degree} to \SI{85}{\degree} to increase the resolution at high incidence angles.
As for the depth, we used $d \in \{1,2,2.5,3,4,5,7,10\}$ meters. Measurements greater than \SI{10}{\meter} were unreliable since the beam becomes larger than the board for high orientations of the board.
To avoid bias due to temperature drift~\cite{Hebert}, sensors were powered at least an hour before recording any data and the metrology room used for the experiment had its temperature controlled at \SI{20}{\celsius}.
Moreover, for each tuple $\{ \theta, d \}$, 96 in total, 45 seconds of measurements were collected to estimate modes and dispersions.

\section{Results} \label{section:Results}
\begin{figure}[t!]
	\centering
	\vspace{-6mm}
	\includegraphics[width=\columnwidth]{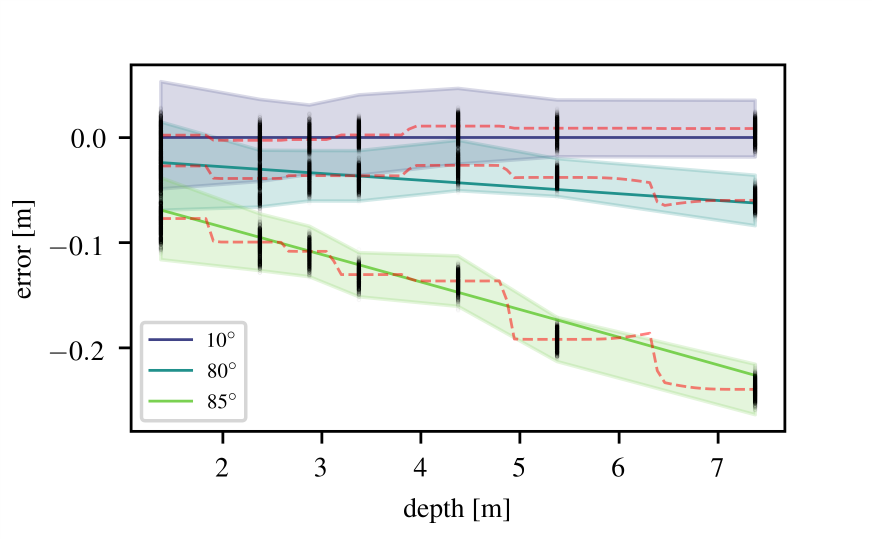}
	\caption{
	Data collected and fit as a function of the depth for several incidence angles for the \LMS.
	Raw data are in black and the filled zones represent the variance evaluated at $3\sigma$.
In red, the non parametric kernel estimation used in \cite{Roca-Pardinas2014} being limited by the sparse data. 
}
	\label{fig:fitLMS}
	\vspace{-4mm}
\end{figure}

An excerpt of the data collected for the \LMS is shown in \autoref{fig:fitLMS}.
We can observe how the depth influences the bias for three incidence angles.
The data show that the main trend of the bias seems to be a linear function of the depth. 
The slope of the curve is nevertheless non-linear and a function of the incidence angle, as we see that the curves differ more between \SI{80}{\degree} and \SI{85}{\degree} than between \SI{10}{\degree} and \SI{80}{\degree}. 
This means that the bias as a function of the incidence angle has a flat region for small angles and grows very quickly for high angles.
\autoref{fig:fitLMS} also shows that the error can be very large for high incidence angles, increasing to more than \SI{20}{\cm} for a distance less than \SI{10}{\m}.
Moreover, some odd measurements were present in the dataset.
For example, at \SI{4}{\meter} in \autoref{fig:fitLMS}, the points deviate a little from the main trend.
A similar observation has been found in the \HDL data.
Our hypothesis is that the sensor electronics adjust themselves when the received intensity becomes too weak. 

\begin{figure*}[htbp]
    \centering
	\vspace{-4mm}
	\includegraphics[width=\linewidth]{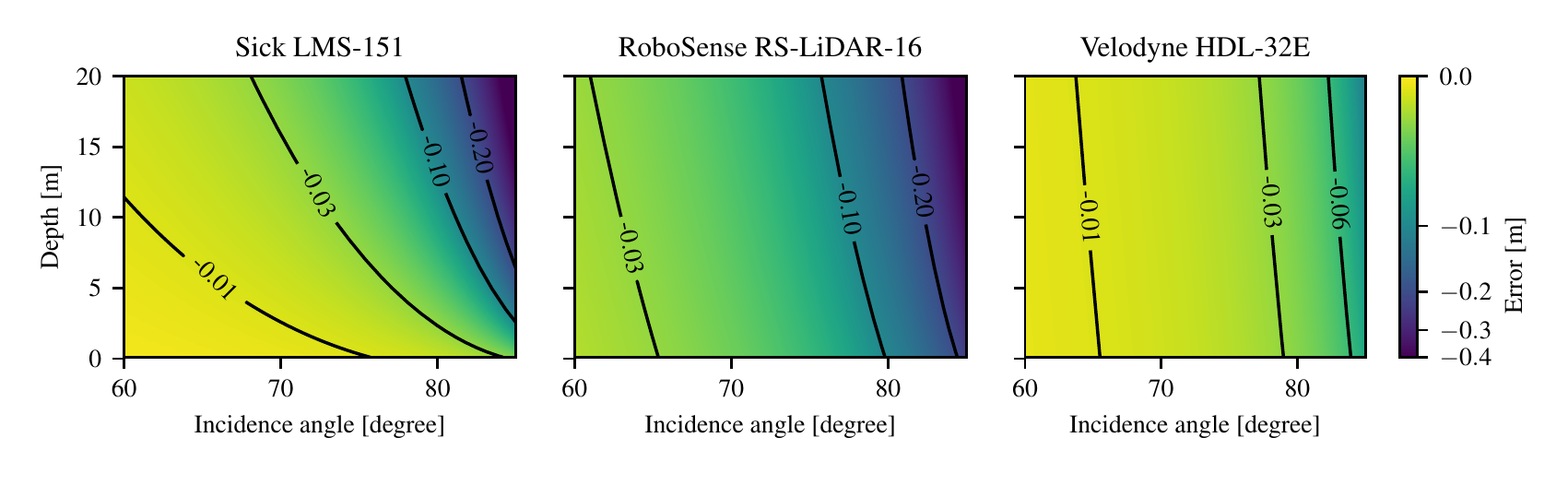}
	\vspace{-11mm}
	\caption{
	Resulting bias estimations using our analytic model.
	Isocurves represent the error as a function of depth and incidence angle for each sensor we evaluated (\LMS, \RS, and \HDL).
Note that the colours representing the bias are on a logarithmic scale.}
	\label{fig:isocurves}
\end{figure*}

We fitted the parameters $(s_1, s_2)$ of \autoref{eq:diff} with these data.
The parameters $I_0$ and $\lambda$ do not influence the metrics. The metrics indeed measure a peak and a ratio of curvature, hence are not influenced by the overall power of the beam. We nevertheless use consistent values for those parameters: $I_0=\SI{0.39}{\watt\per\meter\squared}$ and $ \lambda=\SI{905}{\nano\meter}$.
We experimented different values for the parameter $\tau$ and observed that it also does not influence the metrics as long as it does not become too large. We used $\tau=\SI{50}{\nano\s}$.
Hence, the only parameter which truly impacts the metrics is the aperture half-angle $\alpha$. 
\autoref{tb:params} gives the aperture half-angles as well as the scale parameters found using robust least squares fitting.
%
\begin{table}[t!] 
	\centering
	\caption{Aperture half-angles and scale factors found for each \ac{LIDAR}}
	\begin{tabu}{XXXX}
		\toprule
		Parameters & \LMS  & \RS & \HDL\\ \midrule
		$\alpha$ & \SI{0.43}{\degree} & \SI{0.085}{\degree} & \SI{0.085}{\degree} \\ 
		$s_1$ & $6.08$ & $84.85$ & $10.32$\\  
		$s_2$ & $\SI{3.18e-3}{}$ & $\SI{2.14e-2}{}$ & $\SI{7.08e-3}{}$ \\
		\bottomrule
	\end{tabu}
	\label{tb:params}
	\vspace{-3mm}
\end{table}
The final results of our correction models are plotted in \autoref{fig:isocurves}. 
It is important to note that the bias will always shorten the distance between the sensor and the obstacle.
The plots show that the \LMS and \RS are more prone to large errors than the \HDL. Our hypothesis is that the electronics are different between the sensors, leading to different biases.
The behaviour of the bias also differs between the three sensors: the bias of the \LMS is very variable as a function of the depth, as opposed to the bias of the \HDL which is very stable at constant incidence angles.
\begin{figure*}[htbp] 
	\centering
	\includegraphics[width=\linewidth]{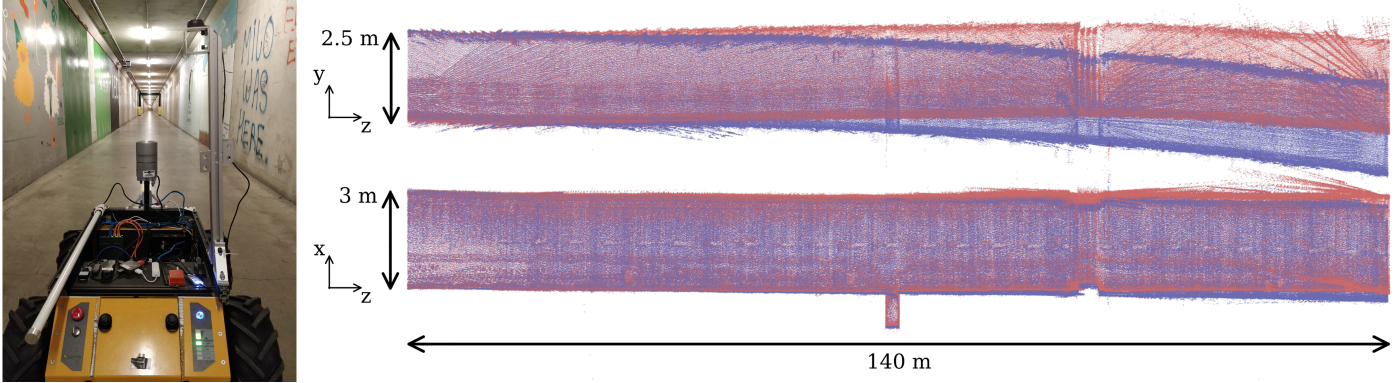}
	\caption{\emph{Left:} \HDL mounted onto the robot in the tunnel.
		\emph{Right:} Side view (\emph{Top}) and top view (\emph{Bottom}) of the tunnel mapped in 3D.
	In blue, the map created with the \HDL.
	In red, the map corrected with our method.
	Although a naive approach was used to compute the surface normals, the map significantly drifts less despite noisy corrections at long ranges (right side of the tunnel).
	Note that the scale on the length of the corridor was changed to highlight the impact.
	The corridor is 47 times longer than its width.}
	\label{fig:beforeAfterSide}
	\vspace{-5mm}
\end{figure*}

We also tested the other models in the literature:
\citet{Lichti2007} chose to remove the points with an incidence angle greater than \SI{65}{\degree}.
This choice is a little extreme as it involves removing all the points of the ceiling and the floor for the \HDL and \RS when mapping a tunnel or corridor, as well as losing the points which are the most constraining in terms of rotation~\cite{Kwok2018}.
\citet{Pfister} modelled the bias with an empirical function $c+bd+ae^{k\theta}$.
This model does not hold, as it assumes that the slope as a function of the depth is not influenced by the incidence angle, which is shown to be false in \autoref{fig:fitLMS}. The slope at \SI{80}{\degree} is greater than the slope at \SI{10}{\degree}.  
Finally,~\citet{Roca-Pardinas2014} used a non parametric kernel estimation, which is a useful fitting tool but is not connected to a physical explanation of the bias.
\autoref{fig:fitLMS} shows the result of the kernel fitting over our data. 
A non parametric kernel estimation will always fit the data but needs more samples than an analytic solution.
Moreover, it is hard to gather experimental data at large distances (e.g., useful for the \HDL and \RS) and the kernel solution will not extrapolate well.

In order to demonstrate that the bias induces a non negligible deformation of the map, we created a map of an underground tunnel of Laval University with the \LMS and \HDL, shown in \autoref{fig:twisted_map} and \autoref{fig:beforeAfterSide}.
Our correction models are implemented in the \texttt{libpointmatcher}, introduced in~\cite{Pomerleau2013} and publicly available to the community%
\footnote{ \href{https://github.com/ethz-asl/libpointmatcher}{https://github.com/ethz-asl/libpointmatcher}}.
Since they rely on a closed-form solution, the computations are lightweight and do not affect the overall performance of the mapping process.
\autoref{fig:twisted_map} shows the result of the correction for a 2D map made with the \LMS.
The map is bending according to our position in the corridor.
This dependence can be explained as the bias of the \LMS is highly dependent on the depth.
For the incidence angle, the points on the farthest side will have a greater distance, hence a greater bias.
Since the bias shortens the distance, the map will bend to the side of the closest wall.
\autoref{fig:beforeAfterSide} shows the result with the \HDL in 3D.
The map before correction (in blue) goes down a few meters, which is significant enough that it can be seen with the bare eye if indeed it was true. 
It does not bend to the left or right, as opposed to the 2D map. It can be explained as the bias does not really depend on the distance (\autoref{fig:isocurves}).
The map bends downward because the floor had higher reflexivity than the ceiling, leading to less perceived points on the floor than on the ceiling (\ie more biased points were contributing to go down rather than up).
One could note some very noisy points on the right side of \autoref{fig:beforeAfterSide}, which were not there before the correction.
Indeed, the estimation of the normals can be noisy at large depths given the sparse configuration of the \HDL, and a small change in the incidence angle at high angles changes drastically the correction.

%
%

\section{Conclusion}
In this paper, we proposed a physical explanation for the bias induced by incidence angles and depths in \ac{LIDAR} measurements.
An approximated closed-form formula was developed and metrics on the return waveform were introduced to cope with the approximations made and the electronics of the sensors.
Three \acp{LIDAR} (\LMS, \HDL, and \RS) were tested in an experimental setup, and biases were found for each sensor.
These biases were demonstrated to increase to \SI{20}{\cm} at a distance less than \SI{10}{\m} for high incidence angles.
Our models were then applied to remove the bias in the measurements of the sensors.
We used the unbiased measurements to create more accurate maps and reduce the drift of the localisation.
Future works should estimate more robustly the normals with an iterative process where the bias is taken into account.
This correction is particularly important for autonomous cars using \acp{LIDAR}, as most of the points on the road have a large incidence angle.

\vspace{-2mm}
\section*{Acknowledgment}
This work was partially supported by the French program WOW! Wide Open to the World in the context of the project I-SITE Clermont CAP 20-25 and the Natural Sciences and Engineering Research Council (NSERC).
We thank the company Robosense for lending their sensor \RS.
We also thank Guy-Jr Montreuil and Christian Larouche from the Department of Geomatics Sciences of Laval University for giving us access to the experimental setup and their help for its construction.

\printbibliography

\end{document}